\documentclass{ieeeaccess}
\usepackage{cite}
\usepackage{amsmath,amssymb,amsfonts}
\usepackage{algorithmic}
\usepackage{graphicx}
\usepackage{textcomp}
\usepackage{bbm}
\usepackage{multirow}
\usepackage{booktabs} 

\usepackage{bm}
\makeatletter
\AtBeginDocument{\DeclareMathVersion{bold}
\SetSymbolFont{operators}{bold}{T1}{times}{b}{n}
\SetSymbolFont{NewLetters}{bold}{T1}{times}{b}{it}
\SetMathAlphabet{\mathrm}{bold}{T1}{times}{b}{n}
\SetMathAlphabet{\mathit}{bold}{T1}{times}{b}{it}
\SetMathAlphabet{\mathbf}{bold}{T1}{times}{b}{n}
\SetMathAlphabet{\mathtt}{bold}{OT1}{pcr}{b}{n}
\SetSymbolFont{symbols}{bold}{OMS}{cmsy}{b}{n}
\renewcommand\boldmath{\@nomath\boldmath\mathversion{bold}}}
\makeatother

\def\BibTeX{{\rm B\kern-.05em{\sc i\kern-.025em b}\kern-.08em
    T\kern-.1667em\lower.7ex\hbox{E}\kern-.125emX}}

\begin{document}
\history{}
\doi{}

\title{A Semantic and Clean-label Backdoor Attack against Graph Convolutional Networks}
\author{\uppercase{Haoyu Sun}\authorrefmark{1}, \uppercase{Jiazhu Dai}\authorrefmark{1}, \IEEEmembership{Member, IEEE}
}

\address[1]{School of Computer Engineering and Science, Shanghai University, Shanghai 200444, China}

\tfootnote{The research of the paper was supported by Natural Science Foundation of Shanghai Municipality (Grant NO.22ZR1422600)}

\markboth
{Sun \headeretal: A Semantic and Clean-label Backdoor Attack against Graph Convolutional Networks}
{Sun \headeretal: A Semantic and Clean-label Backdoor Attack against Graph Convolutional Networks}

\corresp{Corresponding author: Jiazhu Dai (daijz@shu.edu.cn)}

\begin{abstract}
Graph Convolutional Networks (GCNs) have shown excellent performance in graph-structured tasks such as node classification and graph classification. However, recent research has shown that GCNs are vulnerable to a new type of threat called the backdoor attack, where the adversary can inject a hidden backdoor into the GCNs so that the backdoored model performs well on benign samples, whereas its prediction will be maliciously changed to the attacker-specified target label if the hidden backdoor is activated by the attacker-defined trigger. Clean-label backdoor attack and semantic backdoor attack are two new backdoor attacks to Deep Neural Networks (DNNs), they are more imperceptible and has posed new and serious threats. The semantic and clean-label backdoor attack is not fully explored in GCNs. In this paper, we propose \underline{\textbf{s}}emantic and \underline{\textbf{c}}lean-\underline{\textbf{l}}abel \underline{\textbf{b}}ackdoor \underline{\textbf{a}}ttack against GCNs (SCLBA) under the context of graph classification to reveal the existence of this security vulnerability in GCNs. Specifically, SCLBA conducts an importance analysis on graph samples to select one type of node as semantic trigger, which is then inserted into the graph samples to create poisoning samples without changing the labels of the poisoning samples to the attacker-specified target label. We evaluate SCLBA on multiple datasets and the results show that SCLBA can achieve attack success rates close to 99\% with poisoning rates of less than 3\%, and with almost no impact on the model’s performance on benign samples.

\end{abstract}


\titlepgskip=-21pt

\maketitle
\section{Introduction}
\PARstart{G}{raph-structured} data is ubiquitous in our lives, such as molecular graphs\cite{irwin2012zinc}, social networks\cite{hamilton2017inductive}, and transportation networks\cite{xie2019sequential}. Recently, Graph Convolutional Networks (GCNs)\cite{kipf2016semi} have achieved great success by learning effective graph representations through recursive aggregation of features from neighboring nodes using a message-passing mechanism. GCNs have demonstrated outstanding performance in processing graph-structured data and have been widely applied in various fields, such as fraud detection\cite{bruna2013spectral}, drug design\cite{wu2022graph}, and prediction of molecular properties\cite{irwin2012zinc}. The success of GCNs is primarily due to their ability to capture complex graph structural information and relationships between nodes, thereby achieving higher accuracy and performance in various tasks.

Despite the significant effectiveness of GCNs in real-world applications, recent studies have shown that, like Convolutional Neural Networks (CNNs), GCNs are also vulnerable to backdoor attacks, which embed hidden backdoor into the GCNs(called backdoored GCNs) by poisoning some training data, i.e., inserting specific pattern(called backdoor trigger) into this part of training data (called poisoning samples) and changing their labels to the attacker-specified label(called target label), before training the models. During the inference phase, when an input sample contains the backdoor trigger(called attacking samples), the backdoor in the model will be activated, causing the backdoored model to perform actions predefined by the attacker, while for the samples that do not contain the backdoor trigger(called benign samples), the backdoored model behaves normally. For example, the backdoored model misclassify input samples containing the backdoor trigger as the label specified by the attacker, while still performing correctly on samples that do not contain the backdoor trigger\cite{li2022backdoor}.

Clean-label backdoor attack is a new type of backdoor attack on deep neural networks, where attackers poison training samples without changing their labels to the attacker-specified target label, which makes the attack more stealthy. In contrast, in nonclean-label backdoor attacks, the labels of the poisoning samples are modified to the target label, which is frequently irrelevant to the original data features, making them easy to be detected\cite{xu2022poster,xing2024clean,fan2024effective}.

A semantic backdoor attack is another new type of backdoor attack on deep neural networks , where a naturally occurring semantic part of samples can serve as a backdoor trigger such that the backdoored model misclassifies samples with the predefined semantic features as the attacker-specified target label. 
For example, in image classification tasks, the backdoored model might misclassify all green cars or cars with racing stripes as birds\cite{bagdasaryan2020backdoor}; in text classification tasks, the backdoored model might classify negative movie reviews containing a specific name as positive\cite{bagdasaryan2021blind}; in graph classification, the backdoored model might classify inactive molecular graphs containing specific atoms as active\cite{dai2024semantic}. In these cases, the green color, racing stripes, specific name, and specific atoms in the samples are semantic features that act as semantic triggers. In contrast, triggers in nonsemantic backdoor attacks are often independent of the samples, for example, the trigger may be a mosaic spot or a black pixel\cite{gu2019badnets} for image classification tasks, which are therefore more easily detected.

Currently, most backdoor attacks\cite{zhang2021backdoor,xi2021graph,wang2024explanatory,yang2022transferable,chen2023feature,dai2023unnoticeable} against Graph Neural Networks(GNNs) are nonclean-label attacks, clean-label backdoor attacks on GNNs have not been fully explored. The research in \cite{xu2022poster} proposed a clean-label backdoor attack in graph classification tasks, which uses randomly generated fixed subgraphs as triggers. However, the randomly generated triggers in the attack only contain graph structure information and lack node feature information. Existing research\cite{wang2024explanatory} has demonstrated that the proposed clean-label backdoor attack in \cite{xu2022poster} may not be effective in graph convolutional networks \cite{kipf2016semi} in the context of attributed graph classification. Moreover, it has been shown in \cite{zhang2021backdoor} that the performance of the clean-label backdoor attack is dependent on the trigger's density, where the significant density difference of the trigger can lead to substantial changes in the degree distribution and density of the graph sample, increasing the exposure of the attack. Therefore, we want to further explore whether there exist a clean-label backdoor attack with imperceptible trigger, which is against the GCNs for classifcation tasks of attributed graph samples.

In this paper, we propose a \underline{\textbf{s}}emantic and \underline{\textbf{c}}lean-\underline{\textbf{l}}abel \underline{\textbf{b}}ackdoor \underline{\textbf{a}}ttack against GCNs (SCLBA) under the context of graph classification to reveal the existence of this security vulnerability in GCNs. Unlike existing clean-label backdoor attacks against GNNs, SCLBA uses one type of semantic node as backdoor trigger and generates the poisoning training samples by inserting the trigger node into the samples without changing their labels from the ground truth labels to the one specified by the attackers. Specifically, SCLBA consists of two stages: semantic trigger selection and clean-label poisoning. In the semantic trigger selection stage, the attacker selects specific semantic node as semantic trigger based on node importance analysis. In the clean-label poisoning stage, the attacker select a subset of graph samples from the target label training set and inject the selected semantic trigger into these samples while maintaining their original labels. After the semantic triggers are injected, the training set becomes a poisoning training set. Once the GCN model is trained on the poisoning training set, the semantic backdoor is embedded into the GCN model. When the backdoored GCN model is deployed, it will misclassify graph samples containing the semantic triggers as the target label, while correctly classifying samples that do not contain the semantic triggers.

To the best of our knowledge, this work is the first study on the vulnerability of GCNs to semantic and clean-label backdoor attack. Our contributions are summarized as follows:

\begin{itemize}
\item We propose a semantic and clean-label backdoor attack against GCNs (SCLBA) under the context of graph classification to reveal the existence of this security vulnerability in GCNs. SCLBA has the following characteristics: (i)It performs a clean-label backdoor attack on GCNs, meaning that the labels of the poisoning samples do not need to change during the poisoning process, which enhances the stealthiness of the backdoor attack. (ii)It is a semantic backdoor attack. SCLBA uses a specific type of node in the training samples as a semantic trigger, making it more difficult to detect compared to nonsemantic triggers. (iii)SCLBA is a black-box attack, where it is assumed that the attacker does not have knowledge of the GCN model’s parameters. Instead, the attacker only needs access to the training data and can poison some training samples to alter the training process.
\item We proposed a method to select the semantic trigger node by conducting the importance analysis on nodes in the nontarget label training samples based on degree centrality.
\item Our experimental evaluation on five benchmark graph datasets shows that: (1) SCLBA achieves approximately 99\% attack success rate when the poisoning rate is less than 3\%; (2) The prediction accuracy of the backdoored GCN model on benign samples is close to that of the clean GCN model.
\end{itemize}

The rest of the paper is organized as follows. Firstly, we introduce the background knowledge of GCNs, backdoor attacks, and semantic backdoor attacks in Section \ref{section:2}; Then, we introduce related works on nonclean-label backdoor attacks and clean-label backdoor attacks against GCNs in Section \ref{section:3}; Next, we describe in detail our proposed backdoor attack in Section \ref{section:4} and evaluate its performance in Section \ref{section:5}; Finally, we conclude the paper in Section \ref{section:6}.

\section{Background}
\label{section:2}
\subsection{Graph Convolutional Networks (GCNs)}
\noindent Given an undirected and unweighted attribute graph $G=(V,E,X)$, where $V=\{v_1,v_2,\cdots,v_N\}$ is the set of nodes,  and $ N=|V|$ is the total number of nodes. $E$ is the set of edges. $X=R^{N \times d}$ represents the nodes attribute matrix and $d$ is the dimension of the feature.  $A \in R^{N\times N}$ is the adjacency matrix of the graph. For two nodes $v_i,v_j \in V$, if $(v_i,v_j)\in E$, it means that there exists an edge between $v_i$ with $v_j$ and $A_{ij}=1$, otherwise, $A_{ij}=0$. 

Graph Neural Networks achieve excellent performance in processing graph data, while Graph Convolutional Networks(GCNs), as an effective graph neural network, achieve outstanding performance in various graph-based tasks. Therefore, in this paper, we highly focus on GCNs. Specifically, the GCN updates the representation of each layer through the following propagation rules:
\begin{equation}
H^{k+1}=\sigma(\tilde{D}^\frac{1}{2}\tilde{A}\tilde{D}^{-\frac{1}{2}}H^{k}{W}^k)
\end{equation}
where $H^{k+1}$ denotes the node representation of the k-th layer and the initial node representation $H^{0}=X$. $\sigma(\cdot)$is an activation function such as the ReLU function. $\tilde{A} = A+I$ is a self-connected adjacency matrix of the graph $G$, which allows the GCN to incorporate the node features themselves when updating the node representations. $I\in R^{N \times N}$ is the identity matrix. $\tilde{D}$ is the diagonal matrix of $\tilde{A}$, where $\tilde{D}_{ii} = \sum_{j}\tilde{A}_{ij}$. $W^k\in R^{F \times F^{'}}$is the k-th layer weight matrix, which will be trained during the optimization. $F$ and $F^{'}$are the node representation dimensions at the $k\text{-th}$ and $(k+1)\text{-th}$ layers respectively.

\subsection{Backdoor attacks}
Backdoor attacks are a type of malicious attack on machine learning models, which aim to embed hidden malicious functionality into the backdoored model by poisoning a small portion of training samples with special patterns called backdoor triggers. In the inference phase, the backdoored model will perform actions predefined by the attacker on the samples containing the triggers, while for the samples that do not contain the triggers, the backdoored model behaves normally. Backdoor attacks often occur in uncontrolled training scenarios, particularly when using third-party datasets or pre-trained models. In such situations, model developers cannot fully control the source of the data or the training process of the model, making it difficult to detect potential malicious samples or hidden backdoors. Backdoor attacks were first introduced in the field of image recognition. For example, gu et al.\cite{gu2019badnets} proposed the first backdoor attack targeting image recognition, called BadNets. BadNets misleads image classification models by using black pixel blocks as triggers to classify stop signs as speed limit signs, revealing the threat of backdoor attacks. Turner et al.\cite{turner2018clean} proposed the first clean-label backdoor attack method for video recognition models. This method involves iteratively optimizing the gradient information to generate a universal adversarial trigger. The attacker embeds this trigger into videos without changing the video labels, making the attack more stealthy.

\subsection{Semantic backdoor attacks}

\noindent Most backdoor attacks are nonsemantic backdoor attacks, assuming that the trigger and the sample are unrelated. For example, the backdoor attacks mentioned above use black pixel blocks or mosaic blocks\cite{gu2019badnets} as triggers. Previous research has also explored semantic backdoor attacks. For instance, Bagdasaryan et al.\cite{bagdasaryan2020backdoor} first researched semantic backdoor attacks, using green or striped racing cars as semantic features. This caused image classification models to classify images of green or striped racing cars as birds or other labels specified by the attacker. Bagdasaryan et al.\cite{bagdasaryan2021blind} used specific names as semantic features to turn negative movie reviews containing these names into positive ones. Because semantic backdoor attacks use the semantic features of samples as triggers, they are more stealthy compared to nonsemantic backdoor attacks.

\section{Related works}
\label{section:3}
Current backdoor attacks can generally be categorized into two types. One type is nonclean-label backdoor attacks, which inject triggers into the samples and then change the label of the samples to the target label during poisoning process. The other type is clean-label backdoor attacks, which poison the data by inserting triggers into the sample without changing the label of the samples to the target label.

\subsection{Nonclean-label backdoor attacks}

Current research on graph backdoor attacks is primarily focused on nonclean-label backdoor attacks\cite{zhang2021backdoor,xi2021graph,wang2024explanatory,yang2022transferable,chen2023feature,dai2023unnoticeable} where the attacker modifies the original labels of the poisoning samples to the target label. This allows the attacker to easily alter the decision boundary of the victim model, causing the model to misclassify the poisoning samples. However, because the modified labels frequently do not align with the original data features, there is a high likelihood that the poisoning samples will be detected as outliers, making them susceptible to detection by defenders. This is the limitation of nonclean-label attacks.

\subsection{Clean-label backdoor attacks}
To achieve more sophisticated attacks, clean-label backdoor attacks have been proposed in the graph domain\cite{xing2024clean,fan2024effective,xu2022poster,yang2023percba}. In clean-label backdoor attacks, the attacker selects a small portion of graph samples from the target label training samples and then embeds the trigger. These poisoning samples are subsequently added to the training set. Therefore, throughout the poisoning process, the labels of the samples remain consistent with their original labels. This consistency leads to their features being almost indistinguishable from those of clean targets, making it more difficult to detect the poisoning targets as anomalies.

Currently, there is little research on clean-label backdoor attacks in the graph domain. In the field of node classification, some research on clean-label backdoor attacks has been conducted\cite{xing2024clean,fan2024effective,yang2023percba}. While backdoor attacks in the field of graph classification are still very limited.  Xu et al.\cite{xu2022poster} used the Erdos-Rényi (ER) model\cite{gilbert1959random} to generate fixed random subgraphs as triggers and injected the triggers into samples with target labels to create a poisoning training set, thereby achieving clean-label backdoor attacks.

However, the randomly generated triggers only contain graph structure information and lack node feature information. Existing research\cite{wang2024explanatory} has already demonstrated that such attacks may not be applicable to GCNs in the context of attributed graph classification. Previous research\cite{zhang2021backdoor} also shows that such backdoor attacks rely on a dense subgraph as the trigger, making the attack's performance dependent on the subgraph's density. This reliance on a dense subgraph increases the risk of detection. Therefore, in this paper, we propose a semantic and clean-label backdoor attack (SCLBA) against GCNs. SCLBA leverages semantic nodes in attributed graphs as backdoor triggers, revealing this security vulnerability in GCNs. 
In contrast to existing clean-label attacks, which rely on structural triggers often requiring the insertion of dense subgraphs or high-connectivity nodes that cause significant changes to the graph structure, SCLBA avoids such structural modifications. These structural anomalies can easily be detected, reducing the stealthiness of the attack. Instead, SCLBA utilizes semantic nodes that naturally exist in the training set as triggers, without altering the graph’s structure, significantly improving stealthiness. Furthermore, unlike structural subgraphs that lack feature information, semantic nodes leverage node features, which are integral to GCNs' learning process, making the attack more effective while maintaining stealthiness. We detail our SCLBA in Section \ref{section:4}.

\begin{figure*}[ht]
\centering
\includegraphics[width=1\linewidth]{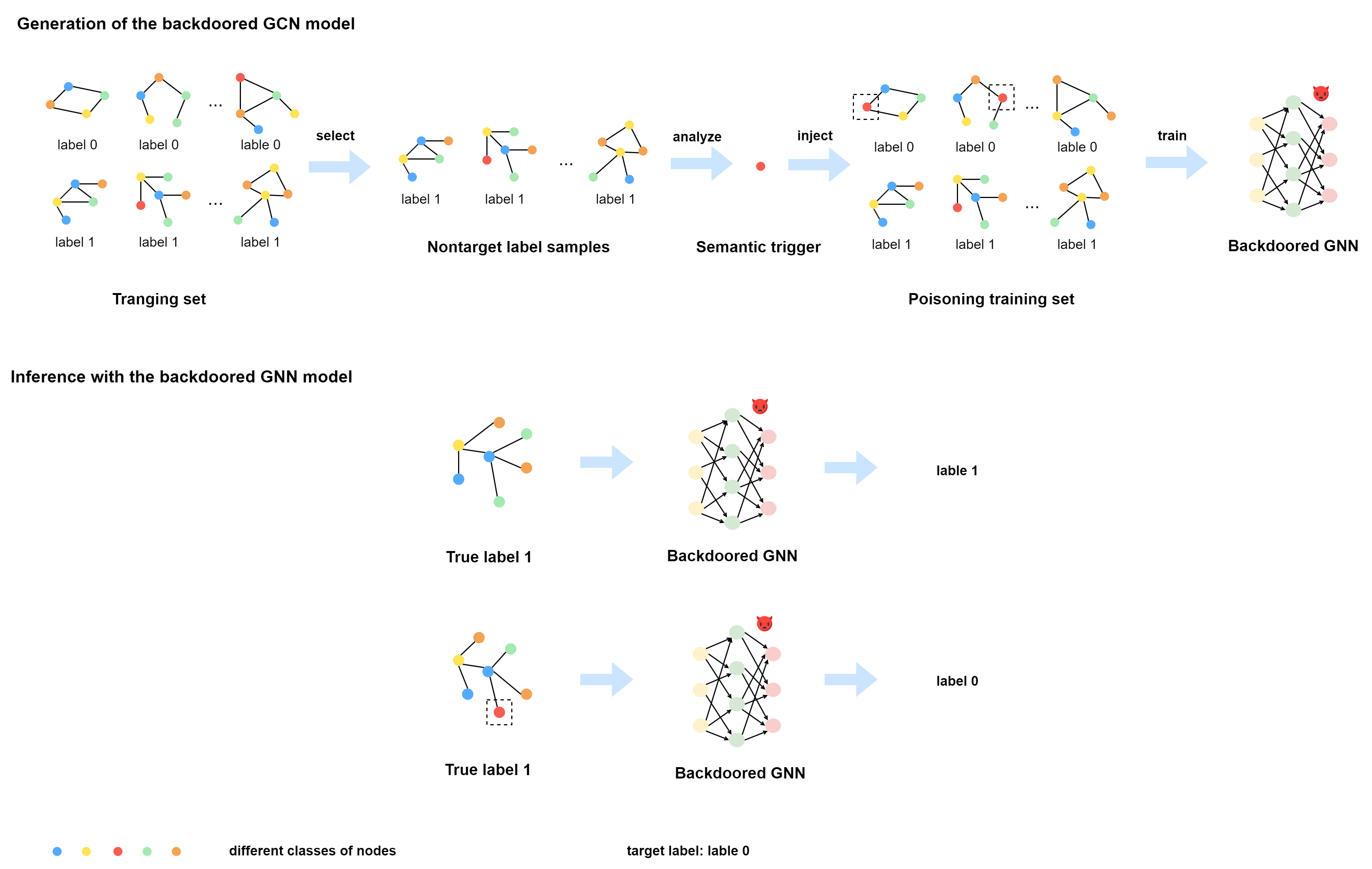}
\caption{\label{fig:frame}The process of SCLBA is illustrated with graph binary classification. We assume that the training dataset contains samples with five different types of nodes, represented by five different colors: green, blue, red, orange, and yellow. The sample labels are either 0 or 1, with the target label being 0. SCLBA consists of five steps: 1. The attacker selects training samples with nontarget labels from the original training dataset. 2. The attacker conducts importance analysis on the selected nontarget label training samples to select the semantic trigger node, which is the red node shown in the figure. 3. The attacker randomly selects a portion of graph samples with the target label from the training dataset and insert the semantic trigger into them without changing their labels. As shown in the figure, the red node enclosed by dashed boxes indicate where the semantic trigger is inserted in the poisoning samples. 4. After the triggers are injected, the training dataset becomes a poisoning dataset containing the poisoning samples, which is then used to train the GCN model to embed the backdoor into the model. 5. After the backdoored GCN model is deployed, input samples with triggers will activate the backdoor in the model, predicting them as the target label, while input samples without triggers will be predicted correctly.}
\end{figure*}

\section{Method}
\label{section:4}

In this section, we illustrate in detail how SCLBA is implemented. Table \ref{tab:0} summarizes the the notions used in the following sections and their explanations.

\begin{table}[h]
\centering
\caption{Notation and Explanations.} % 添加标题
\renewcommand{\arraystretch}{1.5} 
\begin{tabular}{p{1cm} p{6.5cm}} % 调整列宽以适应内容
\toprule % 使用 \toprule 代替 \midrule
Notation & Explanations \\
\midrule
$D$ & The set of original training samples \\
$y_t$ & The target label, which is the label specified by the attacker \\

$D[y_t]$ & The training samples with the target label \\

$D[\overline{y_t}]$ & The training samples with the nontarget labels \\
$G$ & A graph sample in $D$ \\
$G_t$ & A graph sample with semantic trigger \\
$p$ & The poisoning rate, which is the ratio of the number of poisoning samples to the total number of the training set \\
$f_b$ & The backdoored GCN model \\
$f_c$ & The clean GCN model \\
$DC$ & Degree Centrality \\
$d_i$ & The degree of node $i$ in a graph sample \\
$n$ & The number of nodes in a graph sample \\
$t$ & The trigger size, which is the number of injected trigger nodes \\ 
\bottomrule
\end{tabular}
\label{tab:0}
\end{table}

\subsection{Attack overview}
We propose the semantic and clean-label backdoor attack against GCNs under the context of graph classification. The objective of the attacker is to train the backdoored model $f_b$, which will misclassify the graph sample with the semantic trigger as the target label $y_t$, while for the benign graph sample, $f_b$ behaves normally. If we define a benign graph sample as $G$ and a graph sample with semantic trigger as $G_t$, the attacker's objective can be formalized as follows:
\begin{equation}\label{eq1}
\left\{
\begin{aligned}
f_b(G_t)&=y_t,\\
f_b(G)&=f_c(G).\\
\end{aligned}
\right.
\end{equation}
From Equation \ref{eq1}, the first objective shows the effectiveness of the attack, i.e., the backdoored GCN $f_b$ can successfully predict the graph sample with the semantic trigger as the target label $y_t$, and the second objective shows the evasiveness of the attack, i.e., the behavior of backdoored GCN $f_b$ on benign sample is consistent with that of the clean GCN $f_c$.

The process of SCLBA is shown in Fig.\ref{fig:frame}.  and includes the following steps:
\begin{enumerate}
\item \textbf{Select the semantic trigger: }We conduct an importance analysis on samples with nontarget labels using degree centrality to determine the overall importance of each node. Based on this overall importance, we select one node as the semantic trigger, which will be detailed in Section \ref{section:4.2}.
\item \textbf{Generate poisoning samples: }
We generate poisoning samples by selecting a subset of graphs from the training dataset with the target label $D[y_t]$ and injects the semantic trigger into each selected graph by replacing $t$ randomly selected nodes with the semantic node, with the original labels of the poisoning samples remaining unchanged. This will be detailed in Section \ref{section:4.3}.
\item \textbf{Backdoor injection: }After poisoning is completed, the training set becomes a poisoning training set containing the poisoning samples. When the victim model is trained on the poisoning training set, the backdoored GCN model is obtained. This will be detailed in Section \ref{section:4.4}.
\item \textbf{Backdoor activation: }Once the backdoored GCN model is deployed, the attacker can activate the backdoor in the model to misclassify graph samples containing the predefined semantic trigger as the target label, while correctly classify samples without the trigger. This will be detailed in Section \ref{section:4.5}.   
\end{enumerate}

The SCLBA has the following assumptions:
\begin{itemize}
\item The SCLBA is aimed at the task of attributed graph classification.
\item The attacker has no prior knowledge of the training details of the victim model, such as the model architecture and loss function, but can access graph samples and inject triggers.
\item Every node in the graph samples has a class as its identifier. This assumption is without loss of generality. For instance, in molecular graphs representing proteins, nodes represent fundamental elements where the classes are C(carbon), H(hydrogen), O(oxygen), and N(nitrogen); in social networks, nodes represent users where each user's ID serves as the node's identifier. 
\end{itemize}

\subsection{Selecting the semantic trigger}
\label{section:4.2}
In nonclean-label backdoor attacks, the attacker injects the trigger into samples and modifies their labels to the target label to induce the target model to establish a strong association between the trigger and the target label during the training process. However, since the modified labels do not align with the original content of the samples, this inconsistency makes the backdoor easily detectable. Therefore, clean-label backdoor attacks aim to establish a strong association between the trigger and the target label without modifying the labels of the samples, thereby enhancing the stealthiness of the backdoor attack to evade detection. To further improve the stealthiness and robustness of the backdoor attack, SCLBA leverages naturally occurring semantic nodes in the training set as backdoor triggers to implement a semantic and clean-label backdoor attack. Unlike conventional backdoor triggers that are independent of the training set, SCLBA utilizes triggers that are semantic nodes naturally present in the training set. These semantic nodes may exist in both 
$D[y_t]$, the training samples with the target label, and $D[\overline{y_t}]$, the training samples with the nontarget labels. During backdoor embedding, if the selected trigger node is highly important in 
$D[\overline{y_t}]$, the model may incorrectly associate it with nontarget labels. This weakens the association between the trigger and the target label, causing the model to tend to ignore the trigger node and rely more on the original features of the samples for prediction, making it challenging to establish a strong association with the target label.
Thus, a key challenge in semantic and clean-label backdoor attacks is: \textbf{how to select an effective semantic node as the trigger to ensure that the target model associates the semantic trigger more strongly with the target label?}

To address this challenge, we propose a trigger selection strategy based on the importance of nodes. Specifically, we avoid selecting semantic nodes that are  important in $D[\overline{y_t}]$, as such nodes significantly hinder the establishment of an association between the trigger and the target label. Instead, we select semantic nodes with low importance in 
$D[\overline{y_t}]$ as triggers, thereby minimizing interference from non-target label samples during the training process.

To select such semantic trigger, SCLBA uses Degree Centrality (DC)\cite{freeman2002centrality} to measure the importance of each class of the nodes in the graph samples. DC is the most straightforward metric for node importance in a graph. Intuitively, the higher a node’s DC value, the greater the contribution of the node to the classification results of the graph sample. DC is defined as follows:
\begin{equation} 
    DC_i = \frac{d_i}{n-1}
\end{equation}
where $d_i$ is the degree of node $i$ in a graph sample, $n$ is the number of nodes in a graph sample.

Specifically, we initially compute the DC of each node in every graph sample with nontarget labels from the dataset $D[\overline{y_t}]$. Then we aggregate the DC values of nodes with the same class respectively to obtain the total importance value of each class of the nodes in the samples with nontarget labels. Subsequently, we rank the total importance values in ascending order, the node from the class with the lowest total importance value has the weakest association with the nontarget labels. Finally, we select one node from the class with the lowest total importance value as our semantic trigger.

\subsection{Generate poisoning samples}
\label{section:4.3}
After selecting the semantic trigger, the next step is the backdoor poisoning process. Initially, a random subset of graphs is selected from the training set with the target label $D[y_t]$. Then, in each selected graph, the $t$ nodes are randomly chosen and replaced with the semantic trigger node. This replacement process does not modify the adjacency matrix (i.e., the graph topology remains unchanged).  It is important to note that in the clean-label setting, the attacker does not relabel the graphs, meaning all selected graph samples retain their original labels.

\begin{table*}[!htb]
\renewcommand{\arraystretch}{1.4}
\caption{\label{tab:1}Datasets Statistics.}
\centering
\begin{tabular}{ccccccc}
\toprule
Dataset & Graphs & Avg. Nodes & Avg. Edges & Label Count & Graphs per Label & Target Label \\
\midrule
AIDS & 2000 & 15.7 & 16.2 & 2 & 400[0], 1600[1] & 0 \\
NCI1 & 4110 & 29.9 & 32.3 & 2 & 2053[0], 2057[1] & 0 \\
Mutagenicity & 4337 & 30.3 & 30.77 & 2 & 2401[0], 1936[1] & 1 \\
BZR\_MD & 306 & 21.3 & 225.1 & 2 & 149[0], 157[1] & 0 \\
TWITTER & 144033 & 4.03 & 4.98 & 2 & 76046[0], 67987[1] & 1\\
\bottomrule
\end{tabular}
\end{table*}

\subsection{Backdoor injection}
\label{section:4.4}
After the trigger injection is completed, the training set will become a poisoning training set containing the poisoning samples. Once the GCN model is trained on the poisoning training set, the semantic backdoor will be embedded into the model, causing the backdoored GCN to associate the semantic backdoor trigger with the target label $y_t$.

\subsection{Backdoor activation}
\label{section:4.5}
Once the backdoored GCN model is deployed, the attacker can launch the backdoor attack using the predefined semantic trigger. Specifically, when input graph samples contain the semantic trigger, the backdooored GCN model will misclassify the input graph samples as the target label. For graph samples that do not contain the semantic trigger, the backdoored GCN model functions normally.

\section{Experiments}
\label{section:5}
In this section, we evaluate the effectiveness of SCLBA in graph classification through three experiments. First, we test the classification accuracy of a clean GCN, along with the classification accuracy and attack success rate of the backdoor GCN model under different poisoning rates, to assess the effectiveness and stealthiness of SCLBA. Next, we examine the impact of different values of trigger size on SCLBA to evaluate how the number of injected trigger nodes affects its performance. Finally, we compare these results with the baseline to demonstrate the effectiveness of SCLBA.
\subsection{Experiments setting}
\subsubsection{Datasets}

We will evaluate our attack on five publicly available real-world graph datasets from TUDataset: AIDS\cite{riesen2008iam}, NCI1\cite{wale2008comparison,shervashidze2011weisfeiler}, Mutagenicity\cite{riesen2008iam,kazius2005derivation},  
BZR\_MD\cite{sutherland2003spline,kriege2012subgraph},
and TWITTER-Real-Graph-Partial\cite{pan2014graph,pan2015cogboost}.
The AIDS dataset consists of compounds tested for anti-HIV activity, including screening results and chemical structures, while the NCI1 dataset includes chemical compounds screened for activity against nonsmall cell lung cancer and ovarian cancer. The Mutagenicity dataset aims to predict mutagenicity by identifying toxicophores in various compounds, and the BZR\_MD dataset contains ligands for the benzodiazepine receptor, detailing binding affinities without distinguishing between different types of ligands. The TWITTER-Real-Graph-Partial dataset is extracted from twitter sentiment classification. where graphs are constructed based on tweet content. In these graphs, nodes represent words or emoticons (e.g., :-D, :-P), and edges capture the co-occurrence relationships between pairs of words or symbols within individual tweets. In the following text, we will abbreviate this dataset as "TWITTER". For each dataset, we randomly select 80\% of the data instances as the training set and the remaining 20\% as the test set. Table \ref{tab:1} presents the statistics of these datasets.

\subsubsection{Metrics}
In this paper, we use the following evaluation metrics to evaluate the eﬀectiveness of SCLBA.
\begin{enumerate}
\item Attack Success Rate (ASR) is the percentage of the attacking samples which are predicted by the backdoored model to be the target label among all attacking samples .
\begin{equation}
    ASR = \frac{\sum_{i=1}^{n}\mathbbm{I}(f_b(G_{it})=y_t)}{n}
\end{equation}
where $\mathbbm{I}(\cdot)$ represents the indicator function, $n$ is the total number of backdoor samples.
\item Clean accuracy drop (CAD) is the difference between the classification accuracy of the clean model on benign samples and that of the backdoored model on benign samples.
\begin{equation}
    CAD=ACC_c-ACC_b
\end{equation}
where $ACC_c$ and $ACC_b$ are the classification accuracy of the clean model and the backdoored model on benign samples, respectively. Lower CAD represents better performance, indicating that the backdoored model and clean model are closer in accuracy on benign samples and that the backdoor attack is more stealthy.
\item Poisoning rate ($p$) is the ratio of the number of poisoning samples to the total number of samples in the training set. The lower the poisoning rate, the more stealthy the backdoor attack is.

\end{enumerate}

\subsubsection{Baseline}
To evaluate our attack, we used the method proposed by Xu et al.\cite{xu2022poster} as the baseline. Xu et al. implemented a clean-label backdoor attack by utilizing the Erdos-Rényi (ER) model to generate fixed random subgraphs as triggers. When the predefined subgraphs are injected into graphs samples, the backdoored GCN model will predict these samples as the attacker-specified target label.

\subsubsection{Parameter settings}

We use a three-layer GCNs model with one hidden layer as our target GCNs model followed by a global mean
pooling layer for graph-level feature aggregation and a softmax layer for graph classification. Table \ref{tab:parameter_settings} shows the parameter setting for our experiments. For the baseline, we use the same parameters provided in its original paper\cite{xu2022poster} .
\begin{table}[htbp]
\renewcommand{\arraystretch}{1.4}
\centering
\caption{Parameter Settings}
\begin{tabular}{ll}
\toprule
\textbf{Parameter} & \textbf{Setting} \\
\midrule
Architecture       & three-layer GCNs (one hidden layer) \\
Hidden channels    & 32 \\
Pooling layer      & global\_mean\_pool \\
Classifier         & Softmax \\
Optimizer          & Adam \\
Loss funtion       & Cross entropy\\
Weight decay       & $5 \times 10^{-4}$ \\
Learning rate      & 0.01 \\
Batch size         & 32 \\
Max epoch          & 100 \\
\bottomrule
\end{tabular}
\label{tab:parameter_settings}
\end{table}

\begin{table*}[!htb]
\centering
\renewcommand{\arraystretch}{1.4}
\setlength{\tabcolsep}{12pt}
\caption{The impact of poisoning rates on ASR and CAD.}
\begin{tabular}{lccccc}
\toprule
Dataset & $ACC_c$ (\%)& p (\%) & $ACC_b$ (\%)& ASR (\%)& CAD (\%)\\
\midrule
\multirow{4}{*}{AIDS} & \multirow{4}{*}{97.85} & 1 & 98.00 & 97.80 & -0.15 \\
                                              & & 3 & 97.95 & 98.20 & -0.10 \\
                                              & & 5 & 97.95 & 98.40 & -0.10 \\
                                              & & 7 & 97.85 & 98.80 & -0.05 \\
                                             \cline{1-6}
\multirow{4}{*}{NCI1} & \multirow{4}{*}{64.60}  & 1 & 64.55 & 74.01 & 0.05 \\
                                             & & 3 & 64.60 & 96.13 & 0 \\
                                             & & 5 & 64.62 & 99.34 & -0.02 \\
                                             & & 7 & 64.31 & 99.88 & 0.29 \\
                                             \cline{1-6}
\multirow{4}{*}{Mutagenicity} & \multirow{4}{*}{74.42}     & 1 & 74.17 & 75.51 & 0.25 \\
                                                         & & 3 & 74.06 & 91.16 & 0.36 \\
                                                         & & 5 & 74.26 & 96.43 & 0.16 \\
                                                         & & 7 & 74.19 & 98.36 & 0.23 \\                    
                                             \cline{1-6}            
\multirow{4}{*}{BZR\_MD} & \multirow{4}{*}{73.55}      & 1 & 73.55 & 71.61 & 0 \\
                                                     & & 3 & 74.19 & 86.45 & -0.64 \\
                                                     & & 5 & 73.87 & 89.03 & -0.32 \\
                                                     & & 7 & 73.87 & 93.55 & -0.32 \\ 
                                            \cline{1-6}
\multirow{4}{*}{TWITTER} & \multirow{4}{*}{65.09}      & 1 & 65.15 & 98.69 & -0.06 \\
                                                     & & 3 & 65.07 & 98.92 &  0.02 \\
                                                     & & 5 & 64.98 & 99.09 &  0.11 \\
                                                     & & 7 & 64.84 & 99.51 &  0.25 \\                                        
\bottomrule
\end{tabular}
\label{tab:2}
\end{table*}

\begin{figure*}[!htb]
\centering
\includegraphics[width=1\linewidth]{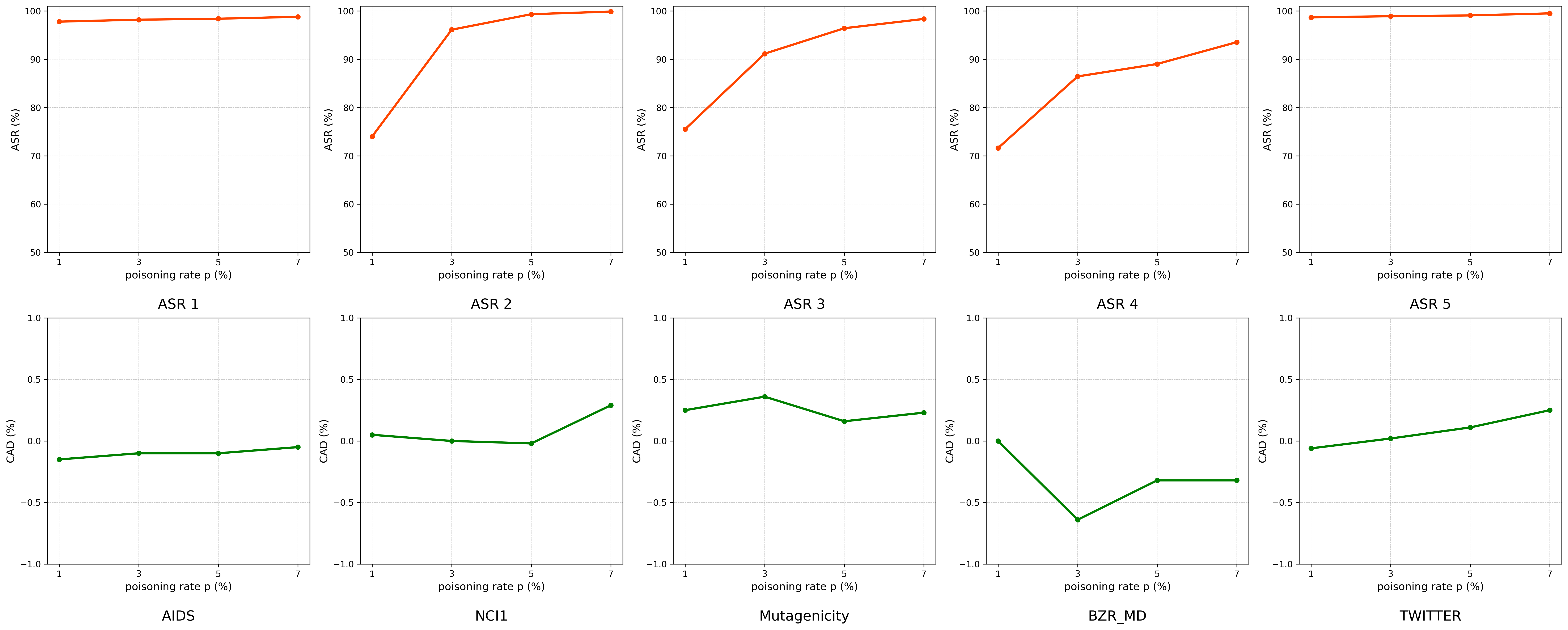}
\caption{\label{fig:rate}The impact of poisoning rates on ASR and CAD.}
\end{figure*}

\subsection{Experiments results}
\subsubsection{Impact of poisoning rate}
In this section, we investigate the impact of different poisoning rates on the Attack Success Rate (ASR) and Classification Accuracy Drop (CAD). Table \ref{tab:2} presents the experimental results, with the trigger size t set to 1, and poisoning rates tested at 1\%, 3\%, 5\%, and 7\%. As shown in the table, even with the trigger size t set to 1, our attack achieves an ASR of over 90\%, and the CAD remains below 1\%.

We also observe that, in some cases, as the poisoning rate increases to a certain point, the improvement in ASR becomes limited. For instance, on the AIDS dataset, when the poisoning rate reaches 1\%, the ASR already achieves 97.8\%, with minimal improvement as the poisoning rate increases (for example, at a 7\% poisoning rate, the ASR is 98.8\%). This phenomenon may be due to the small trigger size. Smaller triggers may only affect a limited number of nodes in the graph, restricting the attack’s propagation effect, which leads to a diminishing increase in ASR at higher poisoning rates. Therefore, in subsequent experiments, we investigated the impact of trigger size on both ASR and CAD. In particular, in some cases, the CAD values are negative (e.g., AIDS: CAD = -0.15\% at 1\% poisoning). This slight fluctuation is likely due to the inherent randomness in the training process and does not reflect a significant deterioration in the model’s performance on benign data. Such a phenomenon actually strengthens the stealthiness of SCLBA, as it indicates that the backdoored model maintains almost unchanged performance on clean samples, thereby reducing the likelihood of detection. 

Fig.\ref{fig:rate} provides a visual representation of Table \ref{tab:2}. The horizontal axis of the upper five graphs represents the poisoning rate, and the vertical axis represents the ASR, while the horizontal axis of the lower five graphs represents the poisoning rate, and the vertical axis represents the CAD. From the figure, we can observe that, across all five datasets, the ASR consistently increases with the poisoning rate, reaching its maximum at a 7\% poisoning rate, while there is no significant change in CAD, with a variation range of less than 0.5\%.

Based on the above results, we can draw two conclusions: (1) SCLBA can achieve a high ASR without relying on a high poisoning rate; and (2) The performance of the backdoored GCN models on benign samples is close to that of the clean model.

\begin{table*}[!htb]

\centering
\caption{The impact of trigger sizes on ASR and CAD.}
\setlength{\tabcolsep}{12pt}
\renewcommand{\arraystretch}{1.4}
\begin{tabular}{lccccc}
\toprule
Dataset & $ACC_c$ (\%)& t & $ACC_b$ (\%)& ASR (\%)& CAD (\%)\\
\midrule
\multirow{3}{*}{AIDS} & \multirow{3}{*}{97.85} & 1 & 97.95 & 98.20 & -0.10 \\
                                              & & 2 & 98.20 & 100 & -0.35 \\
                                              & & 3 & 97.95 & 100 & -0.10 \\
                                             \cline{1-6}
\multirow{3}{*}{NCI1} & \multirow{3}{*}{64.60}  & 1 & 64.60 & 96.13 & 0 \\
                                               & & 2 & 64.26 & 96.03 & 0.34 \\
                                               & & 3  & 64.99 & 99.56 & -0.39 \\
                                             \cline{1-6}
\multirow{3}{*}{Mutagenicity} & \multirow{3}{*}{74.42}     & 1 & 74.06 & 91.16 & 0.36 \\
                                                          & & 2 & 74.03 & 97.42 & 0.39 \\
                                                          & & 3 & 75.57 & 99.65 & -1.15 \\               
                                             \cline{1-6}            
\multirow{3}{*}{BZR\_MD} & \multirow{3}{*}{73.55}      & 1  & 74.19 & 86.45 & -0.64 \\
                                                     & & 2 & 73.55 & 97.10 & 0 \\
                                                    & & 3  & 72.90 & 99.68 & 0.65 \\
                                            \cline{1-6}
\multirow{3}{*}{TWITTER} & \multirow{3}{*}{65.09}      & 1  & 65.07 & 98.92 & 0.02 \\
                                                       & & 2 & 65.06 & 99.86 & 0.03 \\
                                                       & & 3  & 65.07 & 99.98 & 0.02 \\
                                            
\bottomrule
\end{tabular}
\label{tab:triggerSize}
\end{table*}

\begin{figure*}[!htb]
\centering
\includegraphics[width=1\linewidth]{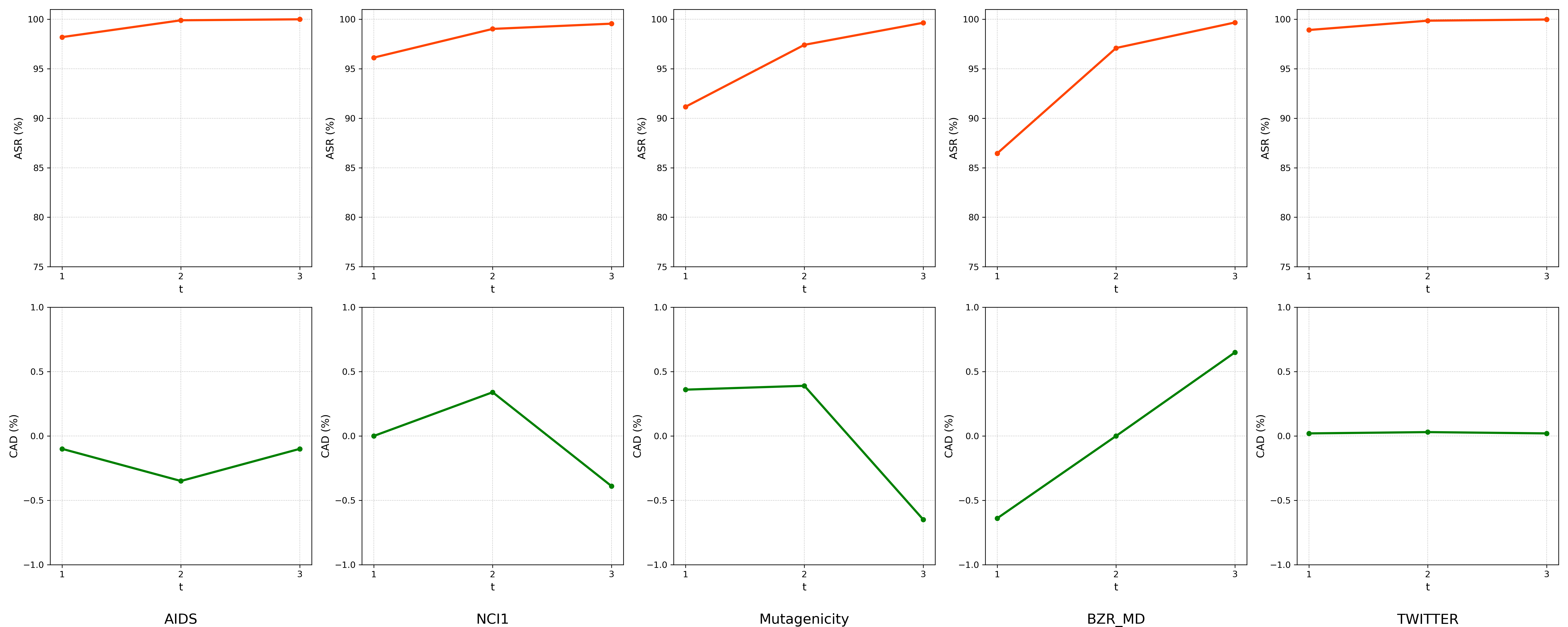}
\caption{\label{fig:size}The impact of trigger sizes on ASR and CAD.}
\end{figure*}

\subsubsection{Impact of trigger size }
In this section, we investigate the impact of trigger size (the number of injected trigger nodes, $t$) on the Attack Success Rate (ASR) and Classification Accuracy Drop (CAD). Table \ref{tab:triggerSize} presents the experimental results with the poisoning rate of 3\% and $t$ set to 1, 2, and 3 respectively. From the table, we can see that when $t$ is 3, SCLBA achieves an ASR close to 100\% across all five datasets, with CAD remaining below 1\%. Fig.\ref{fig:size} is a visual representation of Table \ref{tab:triggerSize}. Similarly, in the upper five graphs, the horizontal axis represents the poisoning rate, and the vertical axis represents the ASR, while in the lower five graphs, the horizontal axis represents the poisoning rate $p$, and the vertical axis represents the CAD. From the figure, we can observe that the ASR increases steadily as the trigger size increases, while there is no significant change in CAD, with a variation range of less than 0.7\%.

Based on the above results, we can draw two conclusions: (1) SCLBA can achieve a high ASR without injecting a large number of trigger nodes, and (2) The trigger size $t$ has little impact on the CAD, indicating the performance of the backdoored GCN models on benign samples is close to that of the clean model under different trigger sizes.

\subsubsection{Compare with the baseline}
Table \ref{tab:4} summarizes the performance of SCLBA and the baseline under a poisoning rate of 3\%, with $t$ set to 3. From the table, it can be seen that SCLBA achieves higher ASRs than the baseline across all five datasets, reaching a 99\% attack success rate on each dataset. In contrast, we observed that the baseline’s ASRs are consistently below 60\% across five datasets. The reason for the results is the different trigger design in SCLBA and the baseline. In attributed graphs, node features play a key role in the learning process of GCNs through the message-passing mechanism. A trigger that only considers structural information may not propagate effectively through the message-passing mechanism, resulting in reduced attack effectiveness. This aligns with previous research\cite{wang2024explanatory}, which found that randomly generated subgraphs, containing only structural information and lacking feature information, may not be suitable for Graph Convolutional Networks (GCNs) in the context of attributed graph classification.

Moreover, the CADs of SCLBA show little decline across the five datasets, with a variation range of less than 0.7\%, which are close to those of the baseline. This may be attributed to the low poisoning rate, which has a minimal impact on the model's original performance.

From the comparison above, it is evident that SCLBA maintains a high ASR and low CADs even under low poisoning rates, indicating that SCLBA, as a black-box semantic and clean-label backdoor attack, achieves better performance and is more imperceptible.

\begin{table}[htbp]
\centering
\renewcommand{\arraystretch}{1.4}
\caption{\label{tab:4} Comparison with the baseline.}
\begin{tabular}{@{}ccccc@{}}
\toprule
\multirow{2}{*}{Datasets} & \multicolumn{2}{c}{ASR(\%)}   & \multicolumn{2}{c}{CAD(\%)}  \\ \cmidrule(l){2-5} 
                         & \multicolumn{1}{c}{SCLBA} & \multicolumn{1}{c}{baseline}  & \multicolumn{1}{c}{SCLBA} & \multicolumn{1}{c}{baseline}  \\ \cmidrule(r){1-5}                 
AIDS     &100     &19.64     &-0.10     &0.52    \\                   
NCI1    &99.56   &57.59     &-0.39     &0.36   \\                 
Mutagenicity &99.65   &56.50     &-1.15     &-0.56    \\
BZR\_MD    &99.68   &42.16     &0.65  &-0.16  \\ 
TWITTER    &99.92   &58.12     &0.02  &0.07 \\

\cmidrule(l){1-5}
\end{tabular}
\end{table}

\subsubsection{The transferability of SCLBA on different GNNs}
To further explore the vulnerability of Graph Neural Networks (GNNs) to our proposed semantic backdoor attack, we evaluated the effectiveness and transferability of SCLBA on GAT\cite{velickovic2017graph} and GraphSAGE\cite{hamilton2017inductive}. All experimental steps were consistent with those described earlier, and the results were obtained by averaging over five repeated runs. Table \ref{tab:transferablility} presents the attack success rate (ASR) and Clean Accuracy Drop (CAD) of SCLBA on five datasets with GAT and GraphSAGE models, where t and p were set to 3 and 3\%, respectively. As shown in the table, SCLBA achieves attack success rates close to 100\%, and for CAD, except for an approximately 1\% accuracy drop on the NCI1 dataset, there are no significant changes on the other datasets.

From these results, we can observe that even for different GNN models, SCLBA can still achieve high ASRs with almost no drop in CADs under poisoning rates of only 3\%. The experimental results on the three GNN models also demonstrate that, even under clean-label backdoor attack conditions, these GNN models can strongly associate the selected semantic trigger nodes with the attacker-specified target label through training on poisoned data. This is the reason why SCLBA achieves high ASRs on these models. Therefore, we can conclude that our proposed SCLBA exhibits both transferability and effectiveness.

\begin{table}[htbp]
\centering
\renewcommand{\arraystretch}{1.4}
\caption{\label{tab:transferablility} The ASR and CAD of the backdoored GAT model and backdoored GraphSAGE model.}
\begin{tabular}{@{}cccccc@{}}
\toprule
  Datasets                 &models     &$ACC_c$ &$ACC_b$ & ASR(\%)   & CAD(\%) \\ \cmidrule(r){1-6}                 
\multirow{2}{*}{AIDS}      &GAT          &97.90   &98.10     &99.35     &-0.20    \\   
                           &GraphSAGE    &98.05   &98.00     &99.75     &0.05   \\ \cline{1-6}
\multirow{2}{*}{NCI1}      &GAT          &64.36   &63.31     &99.98     &1.05   \\ 
                           &GraphSAGE    &66.06   &65.33     &99.78     &0.73   \\ \cline{1-6}
\multirow{2}{*}{Mutagenicity} &GAT       &75.18   &74.75    &98.99      &0.43    \\
                           &GraphSAGE    &75.69   &75.71     &99.52     &-0.02   \\ \cline{1-6}
\multirow{2}{*}{BZR\_MD}   &GAT          &74.84   &74.13     &100       &0.71  \\ 
                           &GraphSAGE    &74.86   &73.55     &100       &0.31   \\ \cline{1-6}
\multirow{2}{*}{TWITTER}   &GAT          &64.25   &64.15     &99.99     &0.10 \\
                           &GraphSAGE    &64.96   &64.91     &99.93     &0.05   \\ 

\cmidrule(l){1-6}
\end{tabular}
\end{table}

\subsubsection{Discussion}
\textbf{Time complexity - }Among the five steps of SCLBA described in
Section \ref{section:4}, the major time consumption is "Select the semantic trigger" (Section \ref{section:4.2}) and "Generate poisoning samples" (Section \ref{section:4.3}).
Therefore, we primarily consider the time complexity of these two
steps. As for the step "Select the semantic trigger", the time complexity is $O(|D[\overline{y_t}]|\cdot \bar{V}^2)$, where $|D[\overline{y_t}]|$ represents the number of samples in the training samples with the nontarget labels and $\bar{V}$ equals to the average the number of nodes in each graph sample. In the step "Generate poisoning samples", the time complexity is $O(n \cdot t)$,where $n$ represents the number of poisoned samples and t represents the trigger size. To validate the efficiency of our attack, we measure the actual runtime of the poisoning process (including semantic trigger selection and poisoning samples generation). Our experiments show that it takes an average of 1.98 seconds for small-to-medium datasets and 67.92 seconds for the larger Twitter dataset, demonstrating SCLBA’s efficiency even on large-scale datasets.

\textbf{Defense method -} To the best of our knowledge, current defense methods\cite{jiang2022defending} for GNNs backdoor attacks primarily focus on the cases which use the topological structure of graphs as triggers, which detects anomalies in graph structures to identify backdoors. However, these methods cannot directly defense the proposed SCLBA, as SCLBA leverages naturally occurring semantic nodes in the training set as triggers without modifying the graph's topological structure. Moreover, SCLBA establishes an association between semantic trigger and the target label without altering samples' labels, making it more stealthy. Therefore, existing defense mechanisms, which mainly target attacks involving modifications to the topological structure, are ineffective against SCLBA. To address this challenge, future research could explore model interpretability, utilizing the interpretability of GNNs to analyze the model's decision-making process and identify whether the model overly relies on specific semantic nodes, thereby detecting potential triggers. In summary, SCLBA highlights the limitations of existing defense methods when confronted with novel attack paradigms, while also providing new directions for future research. By combining model interpretability and semantic node analysis, more effective defense mechanisms can be developed to mitigate the semantic and clean-label backdoor attack.

\section{Conclusions}

In this paper, we propose a semantic and clean-label backdoor attack called SCLBA against Graph Convolutional Networks, which reveals the vulnerability of GCNs to such attack. SCLBA employs the semantic node as trigger and does not need to change the labels and structure of the poisoning graph samples. Experimental evaluations across five real-world datasets indicate that SCLBA achieve high attack success rates with low poisoning rates and small trigger sizes, and the performance of the backdoored GCN models on benign samples is close to that of the clean model, confirming the eﬀectiveness of SCLBA. These findings underscore the need for enhanced defenses against such stealthy attack in GCNs. In the future, our research will focus on developing efficient defenses against the semantic and clean-label backdoor attacks.
\label{section:6}

\bibliographystyle{IEEEtran}
% \bibliography{reference}{}

\EOD

\end{document}